# Longitudinal Sentiment Classification of Reddit Posts

Fabian Nwaoha    Ziyad Gaffar     Ho Joon Chun    Marina Sokolova

University of Ottawa

{fnwao011, cazgaff021, hchun043, sokolova}@uottawa.ca

**Abstract.** We report results of a longitudinal sentiment classification of Reddit posts written by students of four major Canadian universities. We work with the texts of the posts, concentrating on the years 2020-2023. By finely tuning a sentiment threshold to a range of [-0.075,0.075], we successfully built classifiers proficient in categorizing post sentiments into positive and negative categories. Noticeably, our sentiment classification results are consistent across the four university data sets.

## 1 Motivation and Related Work

Our current study focuses on sentiment classification of Reddit posts written by university students. Survey conducted by National College Health Assessment shows that 67% of Canadian university students have grappled with overwhelming anxiety (Abdi, 2023). Those results coincide with earlier results of the Household Pulse Survey which reported that more than 56% of the young adults, 18- 24 years old, have reported the symptoms of depression during the Covid-19 pandemic. It is the highest rate among adult age categories, followed by 48.9% in the 25- 49 years old category (Panchal et al., 2021). However, the above studies did not include sentiment classification of texts posted social networks favored by the student population. We, on other hand, classify the posts into positive, negative, and neutral sentiments. We work with online posts posted on Reddit[1], a social media platform widely used by the students. Specifically, we work with posts from students of four universities: Waterloo, University of Toronto (UofT), McGill, University of British Columbia (UBC).

With over 52 million daily active users (Backlinko, 2023), Reddit is one of the most used social media platforms. Ability to categorize data through different subreddits enhances a focused sentiment analysis (Chen & Sokolova, 2021). The authors use topic modelling and sentiment classification to study positive and negative sentiments in posts united by content of depression. Several recent studies were able to successfully utilize sentiment analysis, albeit focusing on microblogs such as Weibo and X (formerly Twitter). Those studies include (Xie et al,2020), with the focus on public response to COVID-19 on Weibo, a Chinese microblog platform, (Qi & Shabrina, 2023) that compares the effectiveness of lexicon-based and machine learning methods for sentiment analysis of Twitter data and presents a comparative study of different lexicon combinations, while also exploring the importance of feature generation and selection processes for machine learning sentiment classification. Several publications reported on a longitudinal sentiment analysis of Twitter data, e.g. day-to-day sentiment shifts in tweets posted in a three-month period (Valdez et al, 2020). Longitudinal studies of Reddit posts concentrated on mental health and substance use (Alambo et al, 2021; Yang et al, 2023).

Our longitudinal study differs from the above in that i) we work with the text of the posts; ii) we employ advanced analytics tools for in-depth text studies; iii) we use four year-long time intervals: 2020, 2021, 2022, and 2023.

We apply sentiment classification to extracted texts from the Reddit posts. We work with four year-long intervals: 2020, 2021, 2022, and 2023. Those years cover significant changes in public life caused first by COVID-19 pandemic, then easing the restrictions, and finally lifting all the remaining restrictions. The first instance of COVID-19 emerged in January 2020. Consequently, our investigation encompasses the entire period spanning from 2020 to 2022. Given the commencement date of our research, our emphasis is on data gathered from January to November 2023. We collected data from subreddits of Waterloo[2], McGill[3], UBC[4], and UofT[5], four prominent Canadian universities. From those, UofT, McGill, and UBC are the top three research universities in Canada[6]. Waterloo holds #1 in reputational survey and has been ranked among the top three comprehensive universities (i.e., universities that do not have medical schools)[7].

---

[1] https://www.reddit.com/

We employed Logistic Regression and Support Vector Machine (SVM) models to conduct text and sentiment classification. This process aimed to discern the sentiment—whether positive, negative, or neutral—embedded within the discussions. We used the algorithms' implementations by Scikit-Learn[8]. We visualized performance of the classification tasks though Weka, a data mining tool kit[9]. The dataset, already preprocessed and annotated with sentiment labels, was processed with Weka's suite of algorithms. The outcomes derived from Weka's analysis have provided us with a comprehensive machine-learned perspective on the sentiments. We use Word2Vec which is used to create vector representation of texts. Word2Vec is a neural network-based technique used to generate word embeddings, which are numerical representations of words in the form of vectors. The tool was developed by Mikolov et al (2013). We used implementations by Gensim[10], a Python library. We utilized VADER[11] (Valence Aware Dictionary and Sentiment Reasoner) to calculate the presence of negative and positive words. TextBlob[12] computes polarity scores, which can be crucial in understanding the sentimental context of the post. Utilizing TextBlob, we aimed to identify positive and negative sentiments, extracting keywords that carried significant sentimental weight.

## 2 Data Set Construction

We extracted texts with keywords, such as *stress, worried, nervous, uneasy, distress, fear, concern, unmotivated, anxious, anxiety, unease,* and *mental health*. This approach resulted in a dataset of 4,507 entries. A sizable portion of our data was derived with keywords such as stress, worried, and anxiety. Figure 1 shows which keywords yielded the most substantial number of entries and held greater prevalence in the text.

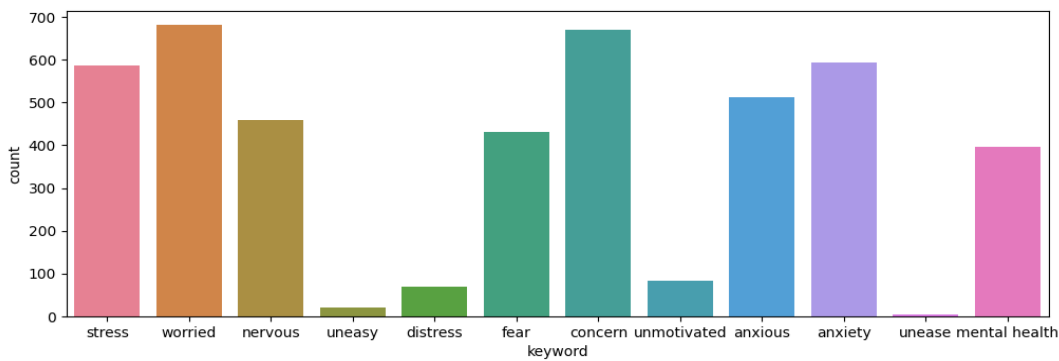

*Figure 1: Keywords and their related data entries*

---

[2] https://www.reddit.com/r/uwaterloo/

[3] https://www.reddit.com/r/mcgill/

[4] https://www.reddit.com/r/UBC/

[5] https://www.reddit.com/r/UofT

[6] Canada's Best Medical Doctoral Universities: Rankings 2024 | Maclean's Education (macleans.ca)

[7] Canada's Best Comprehensive Universities: Rankings 2024 | Maclean's Education (macleans.ca)

[8] https://scikit-learn.org/stable/

[9] https://www.cs.waikato.ac.nz/ml/weka/

[10] https://pypi.org/project/gensim/

[11] https://github.com/cjhutto/vaderSentiment

[12] https://textblob.readthedocs.io/en/dev/

Next, we clean the texts by removing stopwords ('there', 'the', 'is' are examples of stopwords) This step eliminates any commonly occurred words that carry little to no significance in determining sentiment, which leads to the minimization of noise in the dataset. This assists in improving the accuracy and effectiveness of sentiment analysis algorithms in understanding and categorizing sentiments accurately.   We also dropped blank comments that occurred in the data. This led to a decrease in the overall entries from 4,507 to 4,205. Initially, we chose to employ the original dataset with 4,507 entries at the onset of applying sentiment evaluation to the features, aiming to discern performance variations between the two sentiment analysis tools. Subsequently, for the dataset containing 4,205 entries, we employed it specifically during sentiment classification. This decision was driven by the need to eliminate blank comments to reduce potential noise, enhancing machine learning performance and outcomes.

**Feature construction.** The following step was to construct features used by the sentiment analysis algorithms. It begins with the utilization of Word2Vec, and by harnessing the latter, the text data underwent a transformation into meaningful numerical representations. Firstly, we build the Word2Vec model by setting the vector size of the word embeddings to be 100 dimensions, by setting the maximum distance (window) between the current and predicted word within a sentence to 5 words, and finally utilizing the CBOW architecture, which predicts the predicts the target word based on words surrounding it within the defined window. This process takes each word that was tokenized to mapped to a high-dimensional vector space. Then we proceed to derive Word2Vec-based features by computing the average vector representation of words within every textual entry, fusing their semantic essence into cohesive numerical format. The selected features for classification, which will be elaborated upon in the subsequent section, includes the Word2Vec representations of positive, negative, and neutral sentiments, along with the entirety of the text body. Additionally, these features incorporate the total count of positive, negative, and neutral words present in each text body.

**Sentiment evaluation of the features**. In the next stage of the data construction, the above features were evaluated based on the VADER and Textblob sentiment scores.  The procedure goes as follows:

1. We compute both VADER and TextBlob polarity scores for each cleaned text body, which is then placed under a column called 'Vader Sentiment' or 'Textblob Sentiment'.
2. We decide on a threshold value that would be in place to filter out whether the body text or the word is a positive or negative sentiment.
3. After obtaining the best threshold value (more explained under the 'Threshold Value'), utilize it to determine if a text is positive, negative, or neutral.
4. While still using the same threshold value to determine under which sentiment it belongs, we extract the wordlist of each of the sentiment polarities (positive, negative, neutral) by running each word in the cleaned text body through both sentiment tools (VADER and Textblob).
5. Count the total number of words within each of the sentiment wordlist, then placing it under a new column that is specifically for each sentiment polarity (positive, negative, neutral).
6. Utilizing the same Word2Vec model that was used earlier to represent the cleaned text body as word embeddings, we apply it on the wordlist of each sentiment polarity to undergo a transformation towards numerical representations then computing the average vector representations of each word.

**Threshold value**. To move forward towards getting the results for classification, we had to identify the threshold value that would be put in place for the us to determine which cleaned body text or word is positive, negative, or neutral based on the VADER and Textblob polarity scores. So, after replicating numerous datasets, following the preprocessing steps for each dataset but replacing the threshold value for each, we ran each whole dataset through the classification models Logistic Regression and SVM to get their precision, recall and F1-Score, and through these scores we determine which threshold values are the best. Our comparative analysis underscores that Vader demonstrates superior accuracy in assigning sentiment scores compared to TextBlob (Bonta et al, 2019). Thus, we abstain from TextBlob's further use.

We firstly decided to set the threshold value to simply 0.1/-0.1. We initiated the threshold at 0.1/-0.1 based on the observed range of VADER sentiment polarity scores spanning from 1 to -1. This decision aimed to encompass sentiments within a 10 % range above and below the positive/negative score extremes. We aimed to capture stronger positive/negative sentiments while allowing room for some neutral sentiment instances to be included within this range. We duplicated the entire dataset spanning multiple years and universities, systematically varying the threshold values for sentiment classification. Initially, we explored the classification performance by adjusting the threshold incrementally in steps of 0.02 from the initial value of 0.1/-0.1. Through comprehensive classification runs on the complete dataset, the results indicated that a threshold of 0.08 yielded superior outcomes (Table 1).

*Table 1: Macro F-score obtained with the threshold values.*

|  | Threshold values | | | |
|---|---|---|---|---|
|  | **0.075** | **0.080** | **0.100** | **0.120** |
| LR | 0.804 | 0.799 | 0.794 | 0.776 |
| SVM | 0.818 | 0.811 | 0.804 | 0.781 |

**Data sets.** We constructed the data set of 4,205 entries, with 2446 labeled as positive, 1567 - negative, and 192- neutral. The entries are quite balanced in respect to the four years and in respect to the four universities. Table 2 shows the entries split along years and universities.

*Table 2: Distribution of positive, negative, and neutral posts among four years (left) and four universities (right).*

| Year | Pos. | Neg. | Neut. | Total | Universities | Pos. | Neg. | Neut. | Total |
|---|---|---|---|---|---|---|---|---|---|
| **2020** | 447 | 344 | 33 | 824 | **Waterloo** | 514 | 333 | 41 | 888 |
| **2021** | 572 | 384 | 39 | 995 | **UBC** | 650 | 439 | 54 | 1143 |
| **2022** | 652 | 406 | 55 | 1113 | **McGill** | 594 | 360 | 32 | 986 |
| **2023** | 775 | 433 | 65 | 1273 | **UofT** | 688 | 435 | 65 | 1188 |

### 3 Empirical Results of Sentiment Classification

We present results of the deployment of classification models such as Logistic Regression (LR) and Support Vector Machines (SVM). This classification enables us to quantitatively analyze the prevalence and nature of anxiety-related discussions, providing concrete data to understand the mental health landscape among students. Note for the datasets per year, once there are less than 39 Neutral posts, the machine learning models in both WEKA and Scikit-Learn are unable to classify the Neutral sentiment, thus making the Neutral metrics (Precision, Recall, F1-Score) becomes heavily unreliable, and for the rest of the datasets, the number of posts the metrics become unreliable for Neutral is less than 9 Neutral. So, in these following results, any model or value that is missing neutral value has been denoted with a * mark, indicating that the results may be skewed by the lack of Neutral metrics.

For all the results we have obtained, we used 5-fold cross validation to take a comprehensive look at how each model performed across all metrics. We compute Macro metrics. We first report sentiment classification results by year, followed by sentiment classification results by university. Tables 3 and 4 report the results obtained by Scikit-Learn implementations of Logistic Regression and SVM. Recall that the total number of data entries, and distribution of positive, negative, and neutral labels are reported in Table 2. Analyzing these results within the context of the COVID years, we notice that they echo the trends observed across the entire dataset. Notably, we observed minimal differences between performance of the SVM and Logistic Regression models. Classifying the 2023 data set, the Logistic Regression model outperformed the SVM model for the first time in comparison to the other years.

*Table 3: Sentiment classification results for the Reddit posts per the four years.*

|  | Precision | Recall | F1-Score |
|---|---:|---:|---:|
| | **All years** | | |
| **Logistic Regression** | 0.855 | 0.772 | 0.804 |
| **Support Vector Machine** | 0.860 | 0.789 | 0.818 |
| | **2020** | | |
| **Logistic Regression** | 0.561* | 0.581* | 0.569* |
| **Support Vector Machine** | 0.558* | 0.580* | 0.568* |
| | **2021** | | |
| **Logistic Regression** | 0.874 | 0.782 | 0.815 |
| **Support Vector Machine** | 0.865 | 0.787 | 0.814 |
| | **2022** | | |
| **Logistic Regression** | 0.849 | 0.775 | 0.804 |
| **Support Vector Machine** | 0.841 | 0.793 | 0.811 |
| | **2023** | | |
| **Logistic Regression** | 0.861 | 0.797 | 0.823 |
| **Support Vector Machine** | 0.900 | 0.816 | 0.847 |

Our hypothesis for these classification results is that negative posts will yield higher accuracy. We believe this on the notion that negative posts are likely to contain a greater number of distinct words, which grants a higher probability in correct prediction, therefore providing more data for the model's effective training. However, our expectations were defined in terms of negative classification scores surpassing positive ones. Contrary to our hypothesis, across all years. Metrics for positive sentiment consistently outperformed those for negative sentiment by a margin ranging from 3% to as much as 7%. This outcome was rather unexpected for us.

Note that the classification results are relatively stable across the four universities, in F-score as well as in Precision and Recall (Table 4). Such stable results imply a strong generalization ability of our approach.

*Table 4: Sentiment classification of the Reddit posts per university*

|  | Precision | Recall | F1-Score |
|---|---|---|---|
| | Waterloo | | |
| **Logistic Regression** | 0.815 | 0.755 | 0.777 |
| **Support Vector Machine** | 0.864 | 0.784 | 0.808 |
| | McGill | | |
| **Logistic Regression** | 0.863 | 0.739 | 0.774 |
| **Support Vector Machine** | 0.845 | 0.781 | 0.800 |
| | UBC | | |
| **Logistic Regression** | 0.868 | 0.840 | 0.848 |
| **Support Vector Machine** | 0.845 | 0.834 | 0.833 |
| | UofT | | |
| **Logistic Regression** | 0.858 | 0.721 | 0.751 |
| **Support Vector Machine** | 0.836 | 0.735 | 0.760 |

We supplement our classification results by using WEKA visualization of LR and SVM results. Figures 2 – 10 report results obtained from the data sets converted to the WEKA format. Conversion may affect the number of features in the data representation. The number of entries and the labels remain the same as in previous classification experiments. We again first report errors by sentiment classification per year, followed by errors of sentiment classification per university.

In the figures red denotes negative predictions, yellow - neutral, and blue – positive, where the X is for the correct prediction, and the square is for any wrong predictions. Seeing more X indicates that the model was able to predict the class successfully. The plots depict classifiers' proficiency in predicting positive sentiments, as shown by the numerous 'X' markings on the scatter plots. This was the case for both Logistic Regression and SVM. Negative sentiments also show a certain level of correct predictions, but to a lesser extent than positive sentiments. Neutral sentiments have performed worse than the rest of the sentiment classes, which is due to scarcity of the neutral examples. Our analysis indicates a higher rate of correct predictions with the SVM model, demonstrating its enhanced effectiveness. In contrast, while still proficient, Logistic Regression shows a lower accuracy in prediction, underscoring the comparative robustness of SVM in our study.

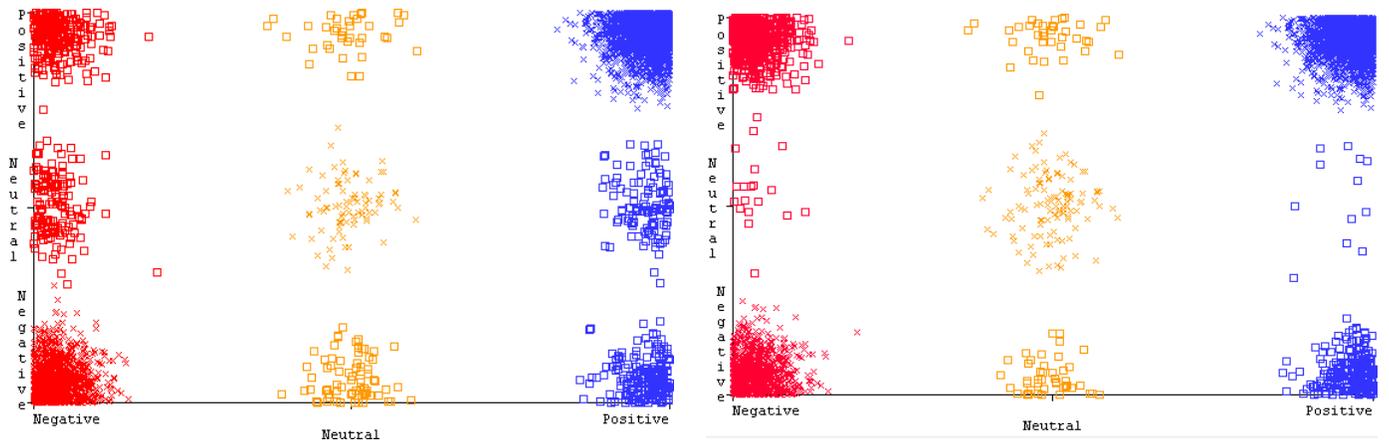

Figure 2: Error Evaluation Scatter Plot of the entire dataset. LR results are to the left, SVM – to the right.

Figures 3-6: Error Evaluation Scatter Plots of the 2020 – 2023 datasets respectively. LR results are to the left, SVM – to the right.

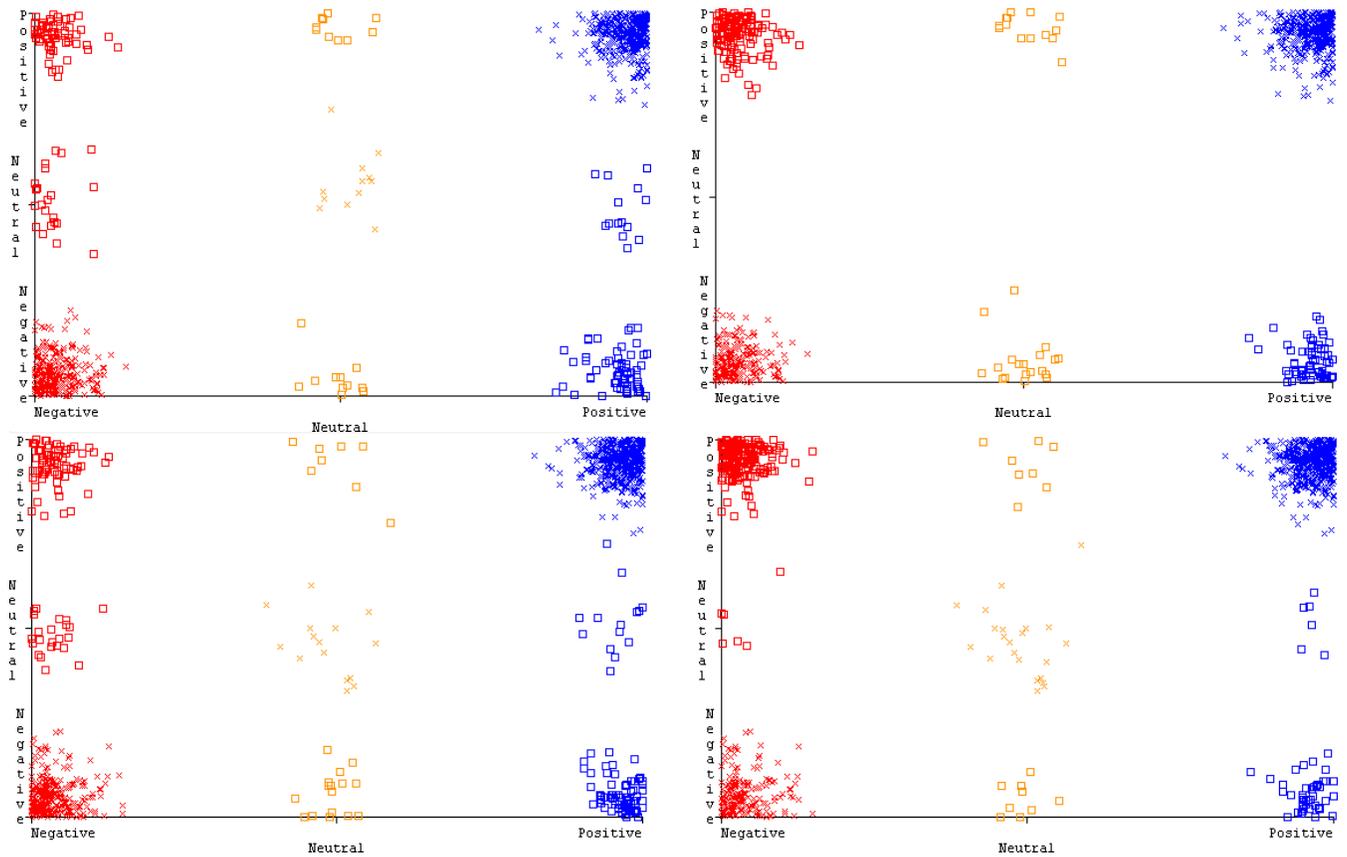

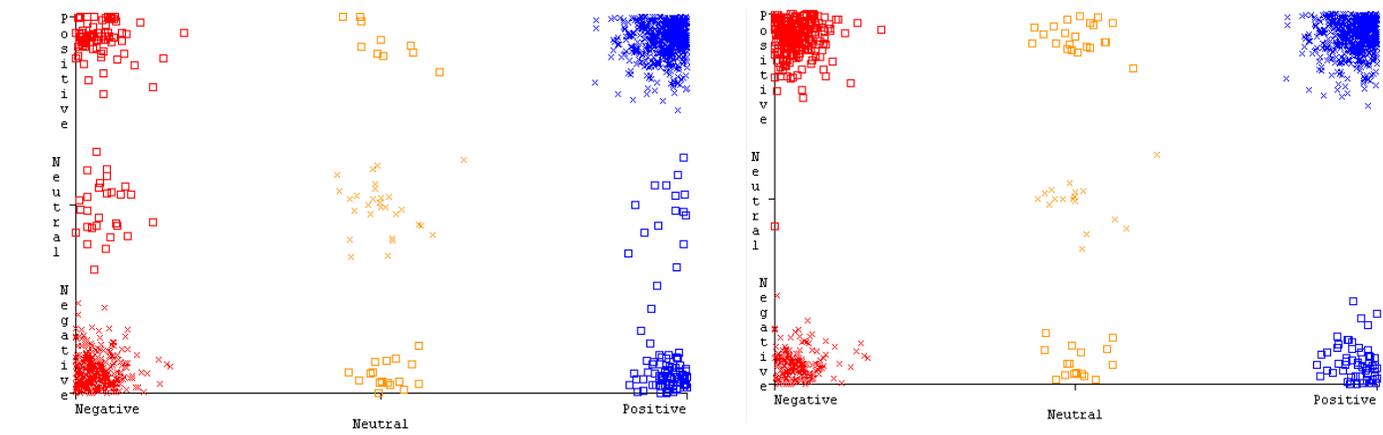
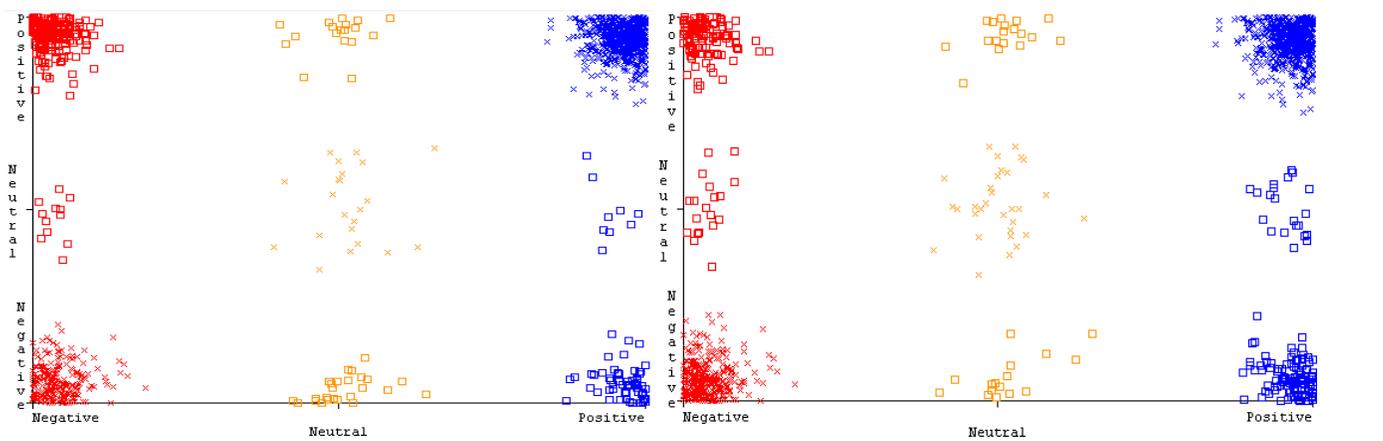

Figures 7 - 10: Classifier Error Evaluation Scatter Plots for Waterloo, McGill, UBC, and UofT respectively; LR- to the left, SVM – to the right.

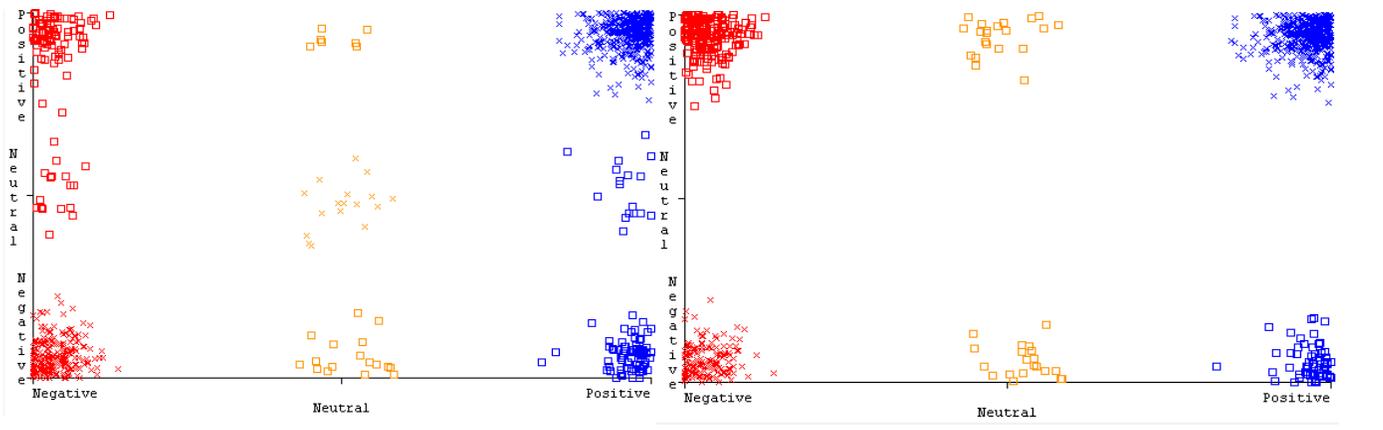

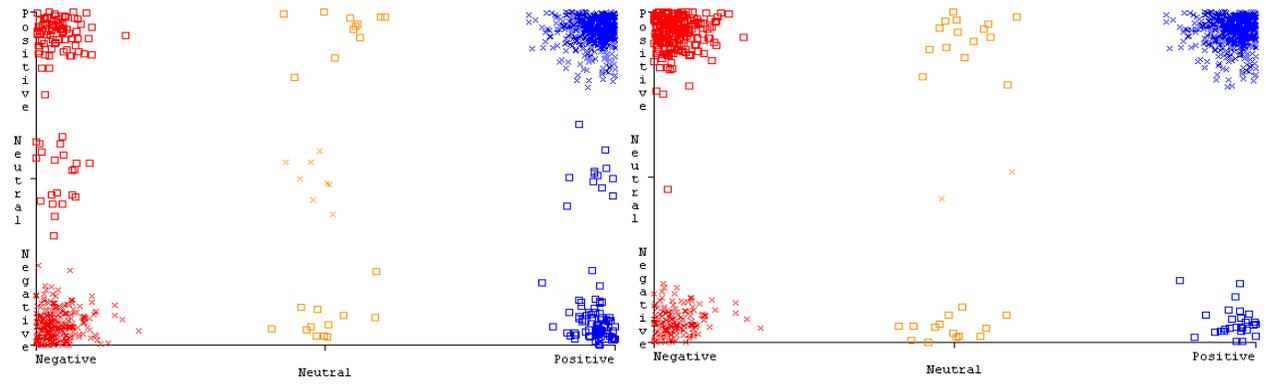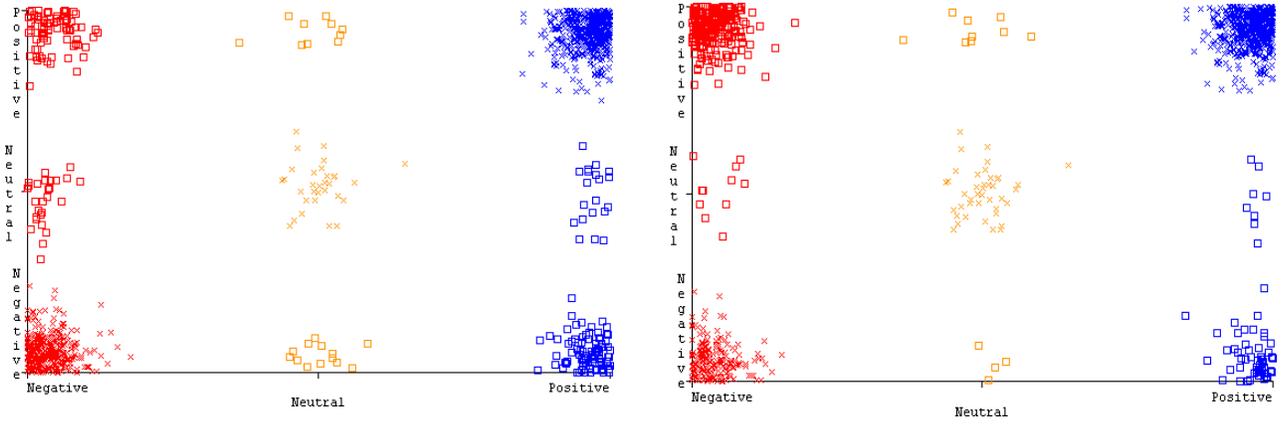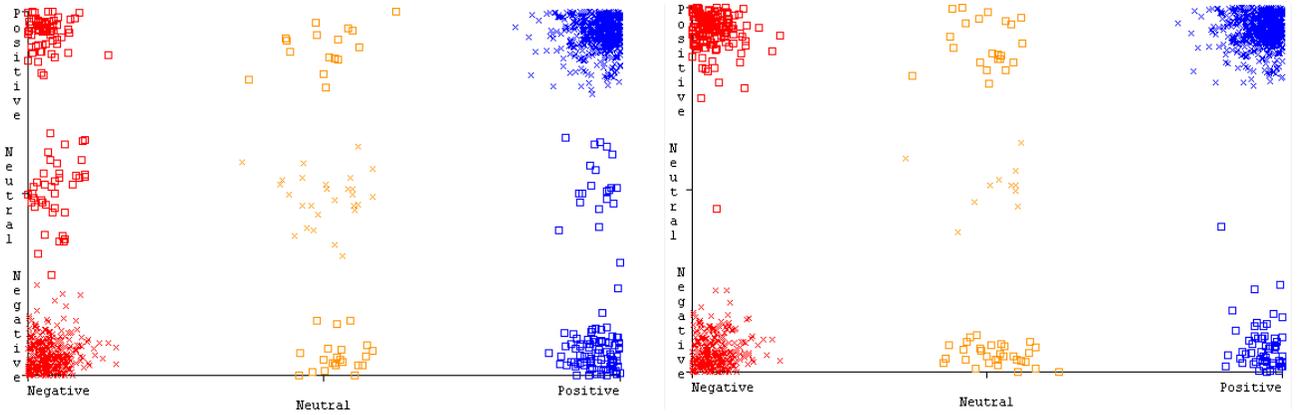

## 4 Result Analytics

The classification results reveal that there was a slight uptick in Precision and Recall for positive posts in 2022-2023, suggesting a potential surge in positive topics during those years (Table 3). Moreover, the dataset itself is predominantly composed of positive topics, contributing to consistently higher Precision and Recall when compared to the outcomes for negative posts.

Noticeably, we obtained consistent sentiment classification results across the data sets from the four universities (Table 4). The consistency implies a generalization power of our results. At the same time, our results are limited in scope. First, we worked with subreddits of predominantly English (language) universities. However, Canada is a bilingual French-English country. Thus, our study omits the students' population that study in French and most probably will write their online messages in French.

Classifying neutral posts proved to be a challenge, both for humans and computers. Adjusting the threshold for what qualified as a neutral term multiple times was necessary to achieve the best possible classification results. After carefully incrementing we achieved the best results with a threshold of 0.075. The scarcity of neutral terms in the dataset reflects the inherent difficulty the classification system faces in accurately categorizing neutrality.

## 5 Conclusions and Future Work

We used sentiment classification to classify texts of posts extracted from subreddits of four prominent Canadian universities (Waterloo, McGill, UofT, and UBC). The data spanned the years 2020-2023. We used year-long intervals for the longitudinal aspect of the study. Our empirical investigation involved a year-by-year analysis, focusing on calculating sentiments for each post and constructing machine learning models to effectively classify sentiments as positive, negative, or neutral.

One hypothesis that held true was the surge in COVID-related mental health and anxiety between 2020-2021, followed by a decline in 2022-2023. This was confirmed by the classification results, where the precision and recall for positive posts was slightly higher in 2022-2023, possibly reflecting the increased occurrence of positive terms during those years.

Our assumption was that the dataset, derived from generally negative keywords, would predominantly consist of negative sentiment posts. However, positive sentiments held a substantial presence in the data. More work should be done to study this phenomenon. We envision that future work can focus on similar longitudinal studies of online posts from bilingual and predominantly French (language) universities in Canada. Improving classification of neutral posts is another important venue of future work.